\documentclass[runningheads]{llncs}
\usepackage{graphicx}
\usepackage{amsmath,amssymb} 
\usepackage{color}
\usepackage{times}
\usepackage{multirow}

\begin{document}
\title{Real-to-Virtual Domain Unification for End-to-End Autonomous Driving} 

\titlerunning{Domain Unification for End-to-End Driving}
%
\author{Luona Yang\inst{1} \and
Xiaodan Liang\inst{1,2} \and
Tairui Wang\inst{2} \and
Eric Xing\inst{1,2}}
%
\authorrunning{L. Yang, X. Liang, T. Wang and E. Xing}
%

\institute{Carnegie Mellon University, Pittsburgh, PA, USA \\
	\email{\{luonay1, xiaodan1, epxing\}@cs.cmu.edu}\\
	\and
Petuum Inc, Pittsburgh, PA, USA \\
\email{tairui.wang@petuum.com}}
\maketitle              
\begin{abstract}
	In the spectrum of vision-based autonomous driving, vanilla end-to-end models are not interpretable and suboptimal in performance, while mediated perception models require additional intermediate representations such as segmentation masks or detection bounding boxes, whose annotation can be prohibitively expensive as we move to a larger scale. More critically, all prior works fail to deal with the notorious domain shift if we were to merge data collected from different sources, which greatly hinders the model generalization ability. In this work, we address the above limitations by taking advantage of virtual data collected from driving simulators, and present DU-drive, an unsupervised real-to-virtual domain unification framework for end-to-end autonomous driving. It first transforms real driving data to its less complex counterpart in the virtual domain, and then predicts vehicle control commands from the generated virtual image. Our framework has three unique advantages: 1) it maps driving data collected from a variety of source distributions into a unified domain, effectively eliminating domain shift; 2) the learned virtual representation is simpler than the input real image and closer in form to the "minimum sufficient statistic" for the prediction task, which relieves the burden of the compression phase while optimizing the information bottleneck tradeoff and leads to superior prediction performance; 3) it takes advantage of annotated virtual data which is unlimited and free to obtain. Extensive experiments on two public driving datasets and two driving simulators demonstrate the performance superiority and interpretive capability of DU-drive.
	
\end{abstract}

\section{Introduction}
\begin{figure}
	\centering
	\includegraphics[width=0.8\linewidth]{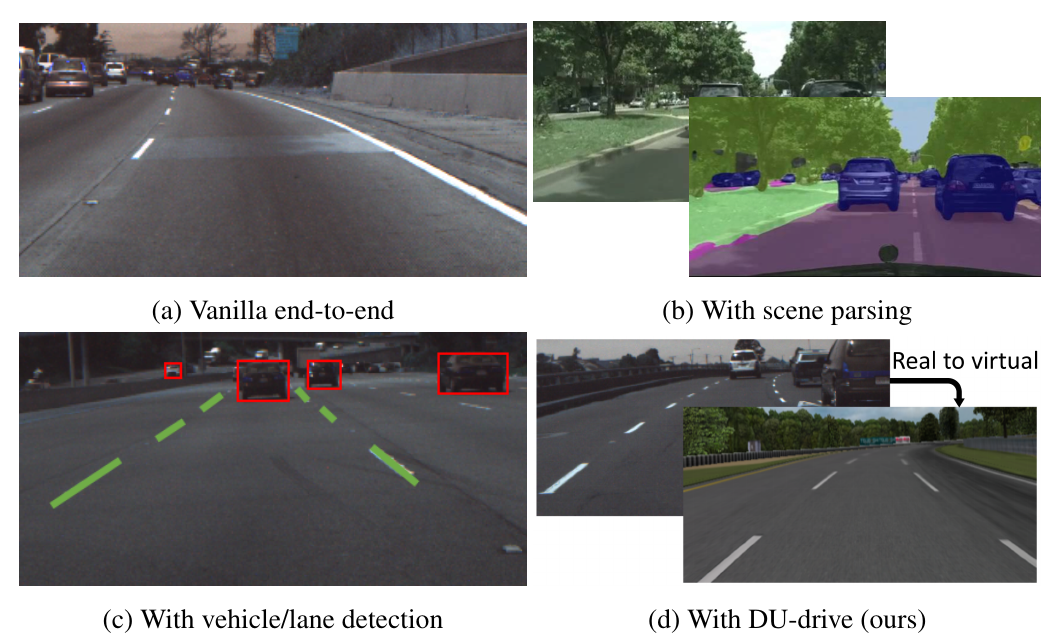}
	\caption{Various methods have been proposed for vision-based driving models. While vanilla end-to-end models (a) are not interpretable and suboptimal in performance, scene parsing (b) or object detection (c) requires expensively annotated data. Our method (d) unifies real images from different datasets into their simpler counterparts in the virtual domain that contains less superfluous details, which boosts the performance of vehicle command prediction task.}
	\label{fig:landscape}
\end{figure}
The development of a vision-based autonomous driving system has been a long-standing research problem~\cite{thorpe1988vision,pomerleau1989alvinn,dickmanns1990integrated,dickmanns1988dynamic}. End-to-end models, among many methods, have attracted much research interest \cite{bojarski2016end,xu2016end,Kim_2017_ICCV} as they optimize all intermediate procedures simultaneously and eliminate the tedious process of feature engineering. \cite{bojarski2016end} trains a convolutional neural network (CNN) to map raw image pixels from a frontal camera to steering commands, which successfully maneuvered the test car in constrained environments. Many attempts have since been made to improve the performance of vanilla end-to-end models by taking advantage of intermediate representations (Figure \ref{fig:landscape}). For example,~\cite{xu2016end} uses semantic segmentation as a side task to improve model performance, while~\cite{chen2015deepdriving} first trains a detector to detect nearby vehicles before making driving decisions. However, the collection of driving data and the annotation of intermediate representation can be prohibitively expensive as we move to a larger scale.

Moreover, raw images of driving scenes are loaded with nuisance details that are not relevant to the prediction task due to the complexity of the real world. For example, a typical human driver will not change his or her behavior according to the shadow of trees on the road, or the view beyond the road boundaries. Such nuisance information could distract the neural network from what is truly important and negatively impact prediction performance. \cite{bojarski2017explaining} visualizes the activation of the neural network and shows that the model not only learns driving-critical information such as lane markings, but also unexpected features such as atypical vehicle classes. \cite{Kim_2017_ICCV} presents results of the driving model's attention map refined by causal filtering, which seems to include rather random attention blobs. 

As pointed out by \cite{tishby2015deep} in the information bottleneck principle, the learning objective for a deep neural network could be formulated as finding the optimal representation that maximally compresses the information in the input while preserving as much information as possible about the output, or in other words, finding an approximate minimal sufficient statistic of the input with respect to the output. Further work \cite{shwartz2017opening} shows that the Stochastic Gradient Descent (SGD) optimization of the neural network has two distinct phases, the fitting phase during which the mutual information of the intermediate layers with the output increases and empirical error drops, and the compression phase during which the mutual information of the intermediate layers with the input decreases and the representation becomes closer in form to the minimum sufficient statistic of the output. They also show that most of the training effort is spent on the compression phase, which is the key to good generalization. It is therefore beneficial for the optimization of the network to have a representation that contains less irrelevant complexity, as it could relieve the burden of the compression phase by giving a better "initialization" of the optimal representation.

More critically, all existing work focuses on a single source of data and does not explicitly deal with generalization to an unseen dataset. As noted by~\cite{torralba2011unbiased}, datasets could have strong built-in biases, and a well-functioning model trained on one dataset will very likely not work so well on another dataset that is collected differently. This phenomenon is known as domain shift, which characterizes the distance in the distribution of inputs and outputs from different domains. While the existing model could be tuned to gradually fit the new domain with the injection of more and more supervised data from the new environment, this could be extremely data inefficient and prohibitively expensive for tasks with diverse application scenarios like autonomous driving.

We propose to tackle the above challenges by taking advantage of virtual data collected from simulators. Our DU-drive system maps real driving images collected under variant conditions into a unified virtual domain, and then predict vehicle command from the generated fake virtual image. Since all real datasets are mapped to the same domain, we could easily extend our model to unseen datasets while taking full advantage of the knowledge learned from existing ones. Moreover, virtual images are "cleaner" as they are less complex and contain less noise, and are thus closer to the "minimal sufficient statistic" of vehicle command prediction task, which is the target representation that the neural network should learn under the information bottleneck framework. Last but not least, our model could take full use of unlimited virtual data and the simulation environment, and a model learned in the virtual environment could be directly applied to a new dataset after unifying it to the virtual domain. Experimental results on two public driving datasets and two driving simulators under supervised and semi-supervised setting, together with analysis on the efficiency of the learned virtual representation compared to raw image input under the information bottleneck framework clearly demonstrate the performance superiority of our method.

\section{Related Work}
\subsection{Vision-based autonomous driving}
Vision-based solutions are believed to be a promising direction for solving autonomous driving due to their low sensor cost and recent developments in computer vision. Since the first successful demonstration in the 1980s~\cite{thorpe1988vision,dickmanns1990integrated,dickmanns1988dynamic}, various methods have been investigated in the spectrum of vision-based driving models, from end-to-end methods to full pipeline methods~\cite{janai2017computer}. The ALVINN system \cite{pomerleau1989alvinn}, first introduced in 1989, is the pioneering work in end-to-end learning for autonomous driving. It shows that an end-to-end model can indeed learn to steer on simple road conditions. The network architecture has since evolved from the small fully-connected network of ALVINN into convolutional networks used by DAVE system \cite{dave} and then deep models used by DAVE-2 system \cite{bojarski2016end}. Intermediate representations such as semantic segmentation masks and attention maps are shown to be helpful to improving the performance \cite{xu2016end,Kim_2017_ICCV}.

Pipeline methods separate the parsing of the scene and the control of the vehicle. \cite{chen2015deepdriving} first trains a vehicle detector to determine the location of adjacent cars and outputs vehicle commands according to a simple control logic.~\cite{huval2015empirical} shows that convolutional neural networks can be used to do real-time lane and vehicle detection. While such methods are more interpretable and controllable, the annotation of intermediate representations can be very expensive. 

Our method takes advantage of an intermediate representation obtained from unsupervised training and therefore improves the performance of vanilla end-to-end driving models without introducing any annotation cost. 

\subsection{Domain Adaption for Visual Data}
Ideally, a model trained for a specific task should be able to generalize to new datasets collected for the same task, yet research has shown that model performance could seriously degrade when the input distribution changes due to the inherent bias introduced in the data collection process~\cite{torralba2011unbiased}. This phenomenon is known as domain shift or dataset bias. In the world of autonomous driving, it is even more critical to have a model that can generalize well to unseen scenarios.

Domain adaption methods attempt to battle domain shift by bridging the gap between the distribution of source data and target data~\cite{bickel2007discriminative,patel2015visual}. Recently, generative adversarial network (GAN) based domain adaption, also known as adversarial adaption, has achieved remarkable results in the field of visual domain adaption.~\cite{tzeng2017adversarial} introduces a framework that subsumes several approaches as special cases~\cite{ganin2015unsupervised,tzeng2015simultaneous,liu2016coupled}. It frames adversarial adaption as training an encoder (generator) that transforms data in the target domain to the source domain at a certain feature level trying to fool the adversarial discriminator, which in turn tries to distinguish the generated data from those sampled from the source domain. The line of work on style transfer~\cite{Zhu_2017_ICCV,Isola_2017_CVPR,johnson2016perceptual} could also be potentially applied to domain adaption at the pixel level.

One subarea especially to our interest is the adaption of virtual data to real data. As the collection of real-world data can be excessively expensive in certain cases, virtual data rendered with computer graphics technologies can come to remedy if we could adapt knowledge learned in the virtual domain to the real domain.~\cite{bousmalis2016unsupervised} proposed a GAN-based model that transforms data from virtual domain to the real domain in the pixel space in an unsupervised manner by utilizing a content-similarity loss to retain annotation.~\cite{Shrivastava_2017_CVPR} uses adversarial training to improve the realism of synthetic images with the help a self-regularization term, a local adversarial loss and a buffer of training images for the discriminator.~\cite{tobin2017domain} randomizes the texture of objects in the robot simulator and trains a visuomotor policy without using any real-world data.~\cite{you2017virtual} trains a driving policy with reinforcement learning in a simulator by transforming virtual images to real images, retaining the scene structure with an adversarial loss on the segmentation mask.

While existing work aims at transforming virtual images to realistic looking images, we argue that doing it the other way around could be more advantageous for learning a driving policy. The transformation from real to virtual is an easier task as it is more manageable to go from complex to simple, and all real datasets could be unified into their simpler counterparts in the virtual domain.
\begin{figure*}
	\begin{center}
		\includegraphics[width=\linewidth,clip,trim={0 0 0 13mm}]{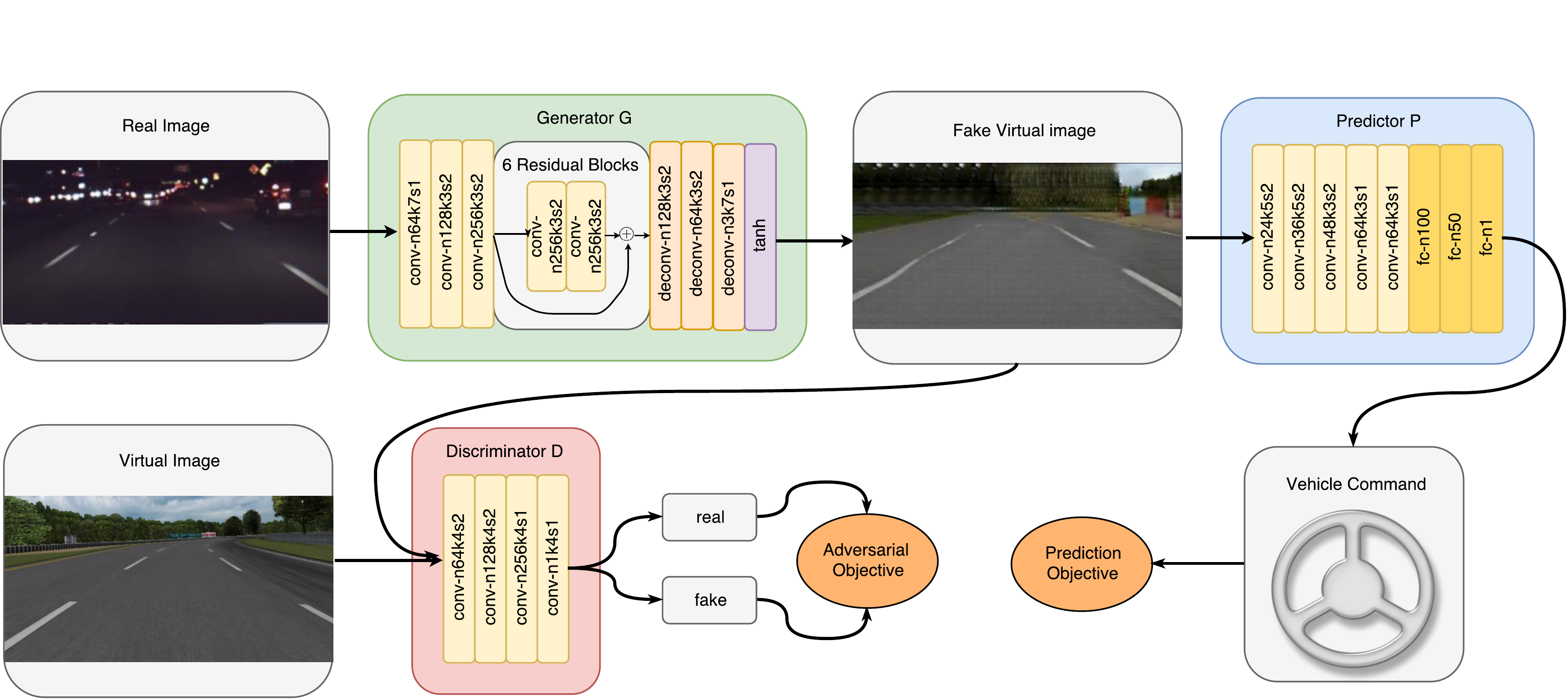}
	\end{center}
	\caption{\small {Model architecture for DU-Drive. The generator network $G$ transforms input real image to fake virtual image, from which vehicle command is predicted by the predictor network $P$. The discriminator network $D$ tries to distinguish the fake virtual images from true virtual images. Both the adversarial objective and the prediction objective drive the generator $G$ to generate the virtual representation that yields the best prediction performance. For simplicity, instance normalization and activation layers after each convolutional/fully connected layer are omitted. (Abbr: n: number of filters, k: kernel size, s: stride size)}}
	\label{fig:archi}
\end{figure*}

\section{Unsupervised Domain Unification}

\subsection{Network Design and Learning Objective}
\textbf{Learning Objective for DU-Drive}
Given a dataset of driving images labeled with vehicle command in the real domain and a similar dataset in the virtual domain, our goal is to transform a real image into the virtual domain and then run prediction algorithm on the transformed fake virtual image. The overall architecture is shown in Fig. \ref{fig:archi}. Our model is closely related to conditional GAN~\cite{mirza2014conditional}, where the generator and discriminator both take a conditional factor as input, yet different in two subtle aspects. One is that in our model, the discriminator does not depend on the conditional factor. The other is that our generator does not take any noise vector as input. Unlike the mapping from a plain virtual image to a rich real image, where there could be multiple feasible solutions, the mapping from a real image to its less complex virtual counterpart should be more constrained and close to unique. Therefore, we could remove the noise term in conventional GANs and use a deterministic generative network as our generator. 

More formally, let $\mathbf{X}^r = \{\mathbf{x}^r_i, \mathbf{y}^r_i\}_{i=1}^{N_r}$ be a labeled dataset with $N^r$ samples in the real domain, and let $\mathbf{X}^v = \{\mathbf{x}^v_i, \mathbf{y}^v_i\}_{i=1}^{N^v}$ be a labeled dataset with $N^v$ samples in the virtual domain, where $\mathbf{x}$ is the frontal image of a driving scene and $\mathbf{y}$ is the corresponding vehicle command. Our DU-drive model consists of a deterministic conditional generator $G(\mathbf{x}^r; \mathbf{\theta}_G) \rightarrow \mathbf{x}^f$, parametrized by $\theta_G$, that maps an image $\mathbf{x}^r \in \mathbf{X}^r$ in the real domain to a fake virtual image $\mathbf{x}^f$, a virtual discriminator $D(\mathbf{x^v}; \theta_D)$ that discriminates whether a image is sampled from true virtual images or from fake virtual images, and a predictor $P(\mathbf{x}^v; \theta_P) \rightarrow y^v$, that maps a virtual image to a vehicle control command. 

The learning objective of DU-drive is:
\begin{align}
\min_{\theta_G, \theta_P}\max_{\theta_D} &\mathcal{L}_d(D, G) + \lambda L_t(P, G), \label{obj}
\end{align}
where $\mathcal{L}_d(D, G)$ is the domain loss, which the generator tries to minimize and the discriminator tries to maximize in the minimax game of GAN. $\mathcal{L}_d(D, G)$ is defined as:
\begin{align}
\mathcal{L}_d(D, G) = &\mathbb{E}_{\mathbf{x}^v}[\log D(\mathbf{x}^v; \theta_D)] +\\
&\mathbb{E}_{\mathbf{x}^r}[\log ( 1 - D(G(\mathbf{x}^r; \theta_G); \theta_D))],
\end{align}
$\mathcal{L}_t (P, G)$ is the task specific objective for predictor and generator, which in this work is the mean square loss between the predicted control command and the ground truth control command, defined as:
\begin{align}
\mathcal{L}_t (P, G)= \mathbb{E}_{\mathbf{x}^r}[ \|P(G(\mathbf{x}^r; \theta_G), \theta_P) - \mathbf{y}^r \|_2^2 ]
\end{align}
$\lambda$ is a hyperparameter that controls the weight of task-specific loss and the domain loss.\\
\textbf{Network Design}\label{network}
For the GAN part of the model, we mostly adopt the network architecture in~\cite{johnson2016perceptual}, which has achieved impressive results in style transfer task. The generator network consists of two convolutional layers with 3x3 kernel and stride size 2, followed by 6 residual blocks. Two deconvolutional layers with stride 1/2 then transform the feature to the same size as the input image. We use instance normalization for all the layers. For the discriminator network, we use a fully convolutional network with convolutional layers of filter size 64, 128, 256 and 1 respectively. Each convolutional layer is followed by instance normalization and Leaky ReLU nonlinearity. We do not use PatchGAN as employed in~\cite{Isola_2017_CVPR} because driving command prediction needs global structure information.

For the predictor network, we adopt the network architecture used in DAVE-2 system\cite{bojarski2016end}, also known as \textit{PilotNet}, as it has achieved decent results in end-to-end driving~\cite{johnson2016perceptual,bojarski2017explaining,bojarski2016end}. The network contains 5 convolutional layers and 4 fully connected layers. The first three convolutional layers have kernel size 5x5 and stride size 3, while the last two layers have kernel size 3x3 and stride size 1. No padding is used. The last convolutional layer is flattened and immediately followed by four fully connected layers with output size 100, 50, 10 and 1 respectively. All layers use ReLU activation. 
\subsection{Learning}
Our goal is to learn a conditional generator that maps a real image into the virtual domain. However, a naive implementation of conditional GAN is insufficient for two reasons. First, the adversarial loss only provides supervision at the level of image distribution and does not guarantee the retention of the label after transformation. Second, conventional GANs are vulnerable to mode collapse, a common pitfall during the optimization of the GAN objective where the distribution of transformed images degenerates. Previous work on adapting virtual image to real image alleviates those problems by introducing a task-specific loss to add additional constraints to the image generated. For example,~\cite{bousmalis2016unsupervised} uses a content similarity loss to enforce that the foreground of the generated image matches with that of the input image.~\cite{Shrivastava_2017_CVPR} employs a self-regularization term that minimizes the image difference between the synthetic and refined images. 

Unfortunately, we cannot take advantage of similar techniques as the "foreground", or the information critical to retaining the label is not obvious for autonomous driving. Instead, we introduce a joint training scheme, where the conditional generator and the predictor are trained simultaneously, so that the supervision from the prediction task gradually drives the generator to convert the input images from the real domain to its corresponding representation in the virtual domain that retains necessary semantics and yields the best prediction performance. 
More formally, our objective in Eq. \ref{obj} can be decomposed into three parts with respect to the three networks $G, P$ and $D$:
\begin{align}
\min_{\theta_G} &\mathcal{L}_d(D, G) + \lambda L_t(P, G), \label{ob1}\\
\min_{\theta_P} &L_t(P, G), \label{ob2}\\
\max_{\theta_D} &\mathcal{L}_d(D, G) \label{ob3}
\end{align}
We omit the weight term $\lambda$ in Equation \ref{ob2}, as it is easy to see that only $\theta_G$ is influenced by both the domain loss and the prediction loss, and we can train $\theta_D, \theta_G$ and $\theta_P$ with respect to the three objectives above independently. We denote $\alpha_P$ as the learning rate for updating $\theta_P$, and $\alpha_{GAN}$ as the learning rate for updating $\theta_D$ and $\theta_G$.

During training, we update $\theta_D, \theta_G$ and $\theta_P$ sequentially by alternately optimizing the above three objectives, so that the generation quality and prediction performance improves hand in hand. 

\subsection{Domain Unification}
Consider the case when we have multiple real datasets $\{\mathbf{x}^{r_1}, \mathbf{y}^{r_1}\}$,...,$\{\mathbf{x}^{r_n}, \mathbf{y}^{r_n}\}$. Due to different data distribution depicted by road appearance, lighting conditions or driving scenes, each dataset belongs to a unique domain which we denote as $D_{r_1}$,...,$D_{r_n}$ respectively. Prior works on end-to-end driving tend to deal with only one domain rather than a more general reasoning system. DU-drive, however, unifies data from different real domains into a single virtual domain and eliminates the notorious domain shift problem.

\begin{figure}[t]
	\begin{center}
		\includegraphics[width=0.8\linewidth]{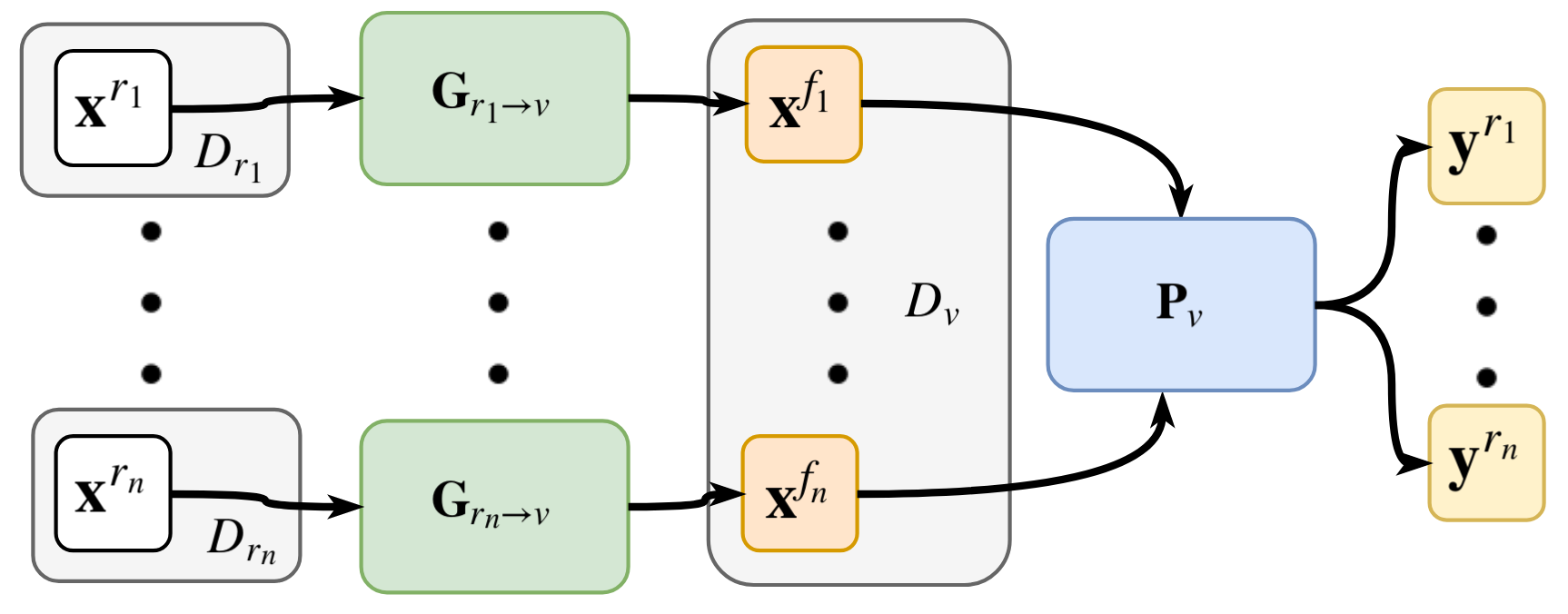}
	\end{center}
	\caption{Domain unification by DU-drive. For each real domain, a generator is trained independently to transform real images to fake virtual images in a unified virtual domain. A single virtual image to vehicle command predictor is trained to do prediction across multiple real domains.}
	\label{fig:du}
\end{figure}

For each real domain $D_{r_i}$, we use our DU-drive model to train a generator that transforms images $\mathbf{x}^{r_i}$ into their counterparts $\mathbf{x}^{f_i}$ in a unified virtual domain $D_v$ (Figure \ref{fig:du}). A global predictor $P_v$ could then be trained to do vehicle command prediction from the transformed virtual images. We fix the generator for each real domain and train the global predictor with labeled data from multiple real domains simultaneously. Same as our training setup for a single domain, we also use \textit{PilotNet} pretrained on virtual data as our initialization for the global predictor. 

\subsection{Connection with Information Bottleneck Principle} \label{IB}
Given a raw image input, what could be a good intermediate representation that could help boost the performance of the prediction task? We try to answer this questions under the information bottleneck framework. 

Formally, let $X$ be the raw image input and $Y$ be the vehicle control command that is to be predicted. The information bottleneck objective of learning for a neural network is to find the optimal representation of $X$ w.r.t. $Y$, which is the minimal sufficient statistic $T(x)$, the simplest sufficient statistic that captures all information about $Y$ in $X$. However, closed form representation for the minimum sufficient statistic does not exist in general, and according to \cite{shwartz2017opening} this objective could be written as a tradeoff between compression of $X$ and prediction of $Y$ formulated in the following form:
\begin{align}
\mathcal{L}[p(t|x)] =  I(X; T) - \beta I(T; Y)
\end{align} 
where $I(X;T)$ denotes the mutual information between the learned representation and input, and $I(T;Y)$ denotes the mutual information between the learned representation and output. This objective is optimized successively for each layer. At the beginning of training, the objective at input layer where $T=X$ could be written as
\begin{align} \label{xr}
L_{\{T=X\}} =& I(X; X) - \beta I(X;Y) \\
=& H(X) - \beta(H(Y) - H(Y|X))\\
=& H(X) - \beta H(Y)\label{ent}
\end{align} 
where Eq. \ref{ent} follows from the fact that $X$ is a sufficient statistic for $Y$.  Now, consider the case when we have an intermediate representation $G(X)$ of $X$. We assume that $G(X)$ is also a sufficient statistic of $Y$, which is reasonable for any meaningful intermediate representation. Then the objective when $T=G(X)$ is
\begin{align}\label{xv}
L_{\{T=G(X)\}} =& I(X; G(X)) - \beta I(G(X);Y) \\ 
=& (H(G(X)) - H(G(X)|X)) - \beta (H(Y) - H(Y|X))\\
=& H(G(X)) - \beta H(Y)
\end{align} 
Subtract Eq. \ref{xv} from Eq. \ref{xr} yields:
\begin{align}
L_{\{T=X\}} - L_{\{T=G(X)\}} = H(X) - H(G(X)) 
\end{align}
This essentially tells us that an intermediate representation with lower entropy could give a better initialization to the information bottleneck objective, which motivates us to transform real images into their simpler virtual counterparts.

\section{Experiments}

\subsection{Data}
\begin{figure*}[th]
	\begin{center}
		\includegraphics[width=\linewidth]{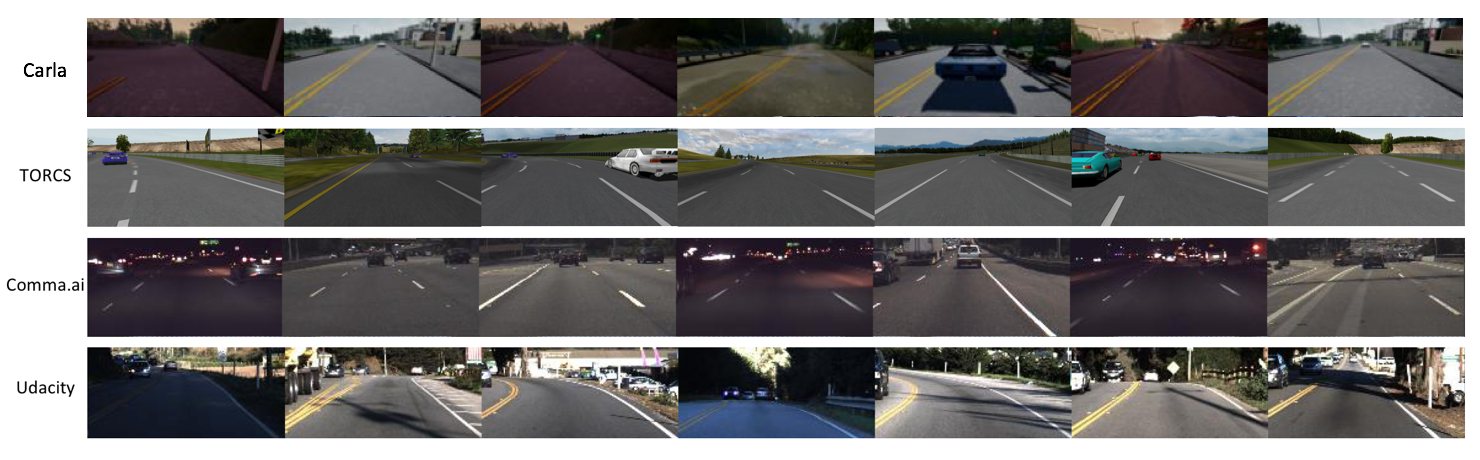}
	\end{center}
	\caption{Sample data used by our work. From top to down: Carla(virtual), TORCS(virtual), Comma.ai(Real), Udacity(real)}
	\label{fig:data}
\end{figure*}

We use TORCS~\cite{wymann2000torcs}, an open-source car racing simulator, and Carla~\cite{Dosovitskiy17}, a recent realistic urban driving simulator as our platform for virtual data collection. Fig. \ref{fig:data} shows samples from both datasets. For TORCS, we construct a virtual dataset by setting up a robot car that follows a simple driving policy as defined in~\cite{chen2015deepdriving} and marking down its frontal camera images and steering commands. We also included twelve traffic cars that follow a simple control logic as defined in~\cite{chen2015deepdriving}, with random noise added to the control commands to encourage varied behaviors. We captured our data on six game tracks with different shapes. To account for the imbalance of right turns and left turns in the virtual data, which could introduce bias in the domain transformation process, we augment our data by flipping the image and negate the steering command. For Carla, we use the training dataset provided by~\cite{Codevilla2018}.

We use two large-scale real-world datasets released by Comma.ai\cite{santana2016learning} and Udacity\cite{udacity} respectively (Table \ref{datasets}). Both datasets are composed of several episodes of driving videos. For Comma.ai dataset, we follow the data reader provided by~\cite{santana2016learning} and filter out data points where the steering wheel angle is greater than 200. For Udacity dataset, we use the official release of training/testing data for challenge II at~\cite{udacity}. Large variance could be observed in lighting/road conditions and roadside views.

\subsection{Preprocessing}
We first crop the input image to 160 x 320 by removing the extra upper part, which is usually background sky that does not change driving behavior. We then resize the image to 80 x 160 and normalize the pixel values to [-1, 1].

Instead of predicting the steering angle command directly, we predict the inverse of the radius as it is more stable and invariant to the geometry of the data capturing car~\cite{Kim_2017_ICCV,bojarski2016end}. The relationship between the inverse turning radius $u_t$ and steering angle $\theta_t$ is characterized by the Ackermann steering geometry:
\begin{align}
\theta_t=u_td_wK_s(1+K_{slip}v_t^2) \label{turnrad}
\end{align}
where $\theta_t$ is the steering command in radius, $u_t$(1/m) is the inverse of the turning radius, $v_t$(m/s) is the vehicle speed at time t. $d_w$(m) stands for the wheelbase, which is the distance between the front and the rear wheel. $K_{slip}$ is the slippery coefficient. $K_s$ is the steering ratio between the turn of the steer and the turn of the wheels. We get $d_w$ and $K_s$ from car specifics released by the respective car manufacturer of the data capturing vehicle, and use the $K_{slip}$ provided by Comma.ai~\cite{santana2016learning}, which is estimated from real data. After predicting $u_t$, we transform it back to $\theta_t$ according to equation \ref{turnrad} and measure the mean absolute error of steering angle prediction.

\begin{table}[h]
	\begin{center}
		\begin{tabular}{|l|c|c|c|}
			\hline
			Dataset & train/test frames & Lighting & size\\
			\hline\hline
			Commai.ai & 345887/32018 & Day/Night & 160 x 320\\
			Udacity & 33808/5279 & Day & 240 x 320\\
			Carla & 657600/74600 & Day/Dawn &88 x 200\\
			TORCS & 30183/3354 & Day & 240 x 320\\
			\hline
		\end{tabular}
	\end{center}
	\caption{Dataset details.}\label{datasets}
\end{table}

\subsection{Training details} 
All the models are implemented in Tensorflow~\cite{abadi2016tensorflow} and trained on an NVIDIA Titan-X GPU. We train all networks with Adam optimizer~\cite{kingma2014adam} and set $\beta_1 = 0.5$. We follow the techniques used in~\cite{Zhu_2017_ICCV} to stabilize the training. First, we use LSGAN~\cite{mao2016multi}, where the conventional GAN objective is replaced by a least-square loss. Thus the loss function becomes
\begin{align}
\mathcal{L}_d(D, G) = &\mathbb{E}_{\mathbf{x}^v}[D(\mathbf{x}^v; \theta_D)^2] +\\
&\mathbb{E}_{\mathbf{x}^r}[( 1 - D(G(\mathbf{x}^r; \theta_G); \theta_D))^2],
\end{align}
Second, we train the discriminator using a buffer of generated images to alleviate model oscillation~\cite{Shrivastava_2017_CVPR}. We use a buffer size of 50. 

In order to take advantage of the labeled data collected from simulators, we initialize the predictor network with a model that is pretrained on virtual images. During pretraining, we set batch size to 2000 and learning rate to 0.01. 

At each step, we sequentially update $\theta_G$, $\theta_P$ and $\theta_D$ with respect to the objective functions in \ref{ob1}, \ref{ob2} and \ref{ob3}. We use a batch size of 60. We set $\alpha_P = 0.0002$, $\alpha_{GAN}=0.00002$, and $\lambda=0.5$ to 1. We train the model for a total of 7 epochs.

After obtaining a real-to-virtual generator for each real domain, we could fix the generator and train a global predictor with all real datasets. We initialize the global predictor with \textit{PilotNet} pretrained on virtual data, and use a learning rate of 0.001 and a batch size of 2000 for training.

\subsection{Metrics and baselines}

We evaluate the effectiveness of our model in terms of the quality of generated images in the virtual domain and the mean absolute error of steering angle prediction. We compare the performance of DU-drive with the following baselines. To ensure fairness, we use the same architecture for the predictor network as described in section \ref{network}.
\begin{itemize}
	\item \textbf{Vanilla end-to-end model (\textit{PilotNet})} proposed by~\cite{bojarski2016end} maps a real driving image directly to the steering command. 
	\item \textbf{Finetune from virtual data} We first train a predictor with virtual data only, then finetune it with the real dataset.
	\item \textbf{Conditional GAN} A naive implementation of conditional GAN (cGAN)~\cite{mirza2014conditional} uses a generator $G$ to transform an image $x$ from the real domain to an image $G(x)$ in the virtual domain. A discriminative network $D$ is set up to discriminate $G(x)$ from $y$ sampled from the virtual domain while $G$ tries to fool the discriminator. No additional supervision is provided other than the adversarial objective. We also train a \textit{PilotNet} to predict steering angle from the fake virtual image generated by cGAN.
	\item \textbf{\textit{PilotNet} joint training} We also directly train a \textit{PilotNet} with two labeled real datasets simultaneously.
\end{itemize}

\subsection{Quantitative Results and Comparisons}
We compare the performance of steering command prediction for a single real domain of our DU-drive(single) model with the plain end-to-end model (\textit{PilotNet}), finetuning from virtual data and conditional GAN without joint training (Table \ref{quant}). Both DU-drive(single) and finetuning from virtual data performs better than the plain end-to-end model, which verifies the effectiveness of leveraging annotated virtual data. DU-drive(single) outperforms finetuning by 12\%/20\% using TORCS virtual data and 11\%/41\% using Carla virtual data for Comma.ai/Udacity dataset respectively, despite using the same training data and prediction network. This verifies the superiority of transforming complex real images into their simpler counterparts in the virtual domain for driving command prediction task. Conditional GAN without joint training does not perform well as adversarial objective itself is not enough to ensure the preservation of label. DU-drive runs at 89.2 fps when tested on a Titan-X GPU.
\begin{table}[h]
	\begin{center}
		\begin{tabular}{|c|c|c|c|c|c|}
			\hline
			\multicolumn{2}{|c|}{Simulator} &  \multicolumn{2}{|c|}{TORCS} &  \multicolumn{2}{|c|}{Carla}\\
			\hline
			Dataset & Model & MAE & SD & MAE & SD\\
			\hline\hline
			\multirow{6}{*}{Udacity} & 
			\textit{PilotNet}\cite{bojarski2016end} & 6.018  & 7.613&6.018&7.613\\
			&Finetune TORCS& 5.808 & 7.721&6.053&8.041\\
			&cGAN~\cite{mirza2014conditional} & 5.921 & 6.896 &4.925&7.100\\
			&\textit{PilotNet} joint training & 15.040 & 27.636& 15.040 & 27.636\\
			&DU-Drive(single) & 4.558 & \textbf{5.356}  &\textbf{3.571}&4.958\\
			&DU-Drive(unified) & \textbf{4.521} & 5.558&3.808&\textbf{4.650}\\
			\hline
			\multirow{6}{*}{Comma.ai} & \textit{PilotNet}\cite{bojarski2016end} & 1.208 & 1.472 & 1.208 & 1.472	\\
			&Finetune TORCS & 1.203 & 1.500 &1.196 & 1.473\\
			&cGAN~\cite{mirza2014conditional}& 1.215 & 1.405 & 1.206 & 1.404\\
			&\textit{PilotNet} joint training & 5.988 & 11.670 & 5.988 & 11.670\\
			&DU-Drive(single) & \textbf{1.061} & 1.319 & \textbf{1.068}&\textbf{1.337}\\
			&DU-Drive(unified) & 1.079 & \textbf{1.270} & 	1.174&1.460\\
			\hline
		\end{tabular}
	\end{center}
	\caption{Mean absolute error (MAE) and standard deviation (SD) for steering angle prediction task. DU-drive clearly outperforms all baseline methods.}\label{quant}
\end{table}

\subsection{Information Bottleneck Analysis of Virtual Representation}
As shown in Table~\ref{quant}, transforming real images to the virtual domain using our DU-drive model gives superior performance even with the same training data and predictor network. We attribute this to the fact that virtual images are more homogeneous and contains less complexity that is not related to the prediction task. As shown in Figure \ref{fig:dudrive}, superfluous details including views beyond the road and changeable lighting conditions are unified into a clean, homogenious background, while cues critical for steering angle prediction like lane markings are preserved. In the languange of information bottleneck theory, this corresponds to a representation that is closer to the optimal minimum sufficient statistic than the raw image with respect to the prediction task.

\begin{figure}
	\centering
	\includegraphics[width=0.7\textwidth, trim={0 6mm 0 8mm}, clip]{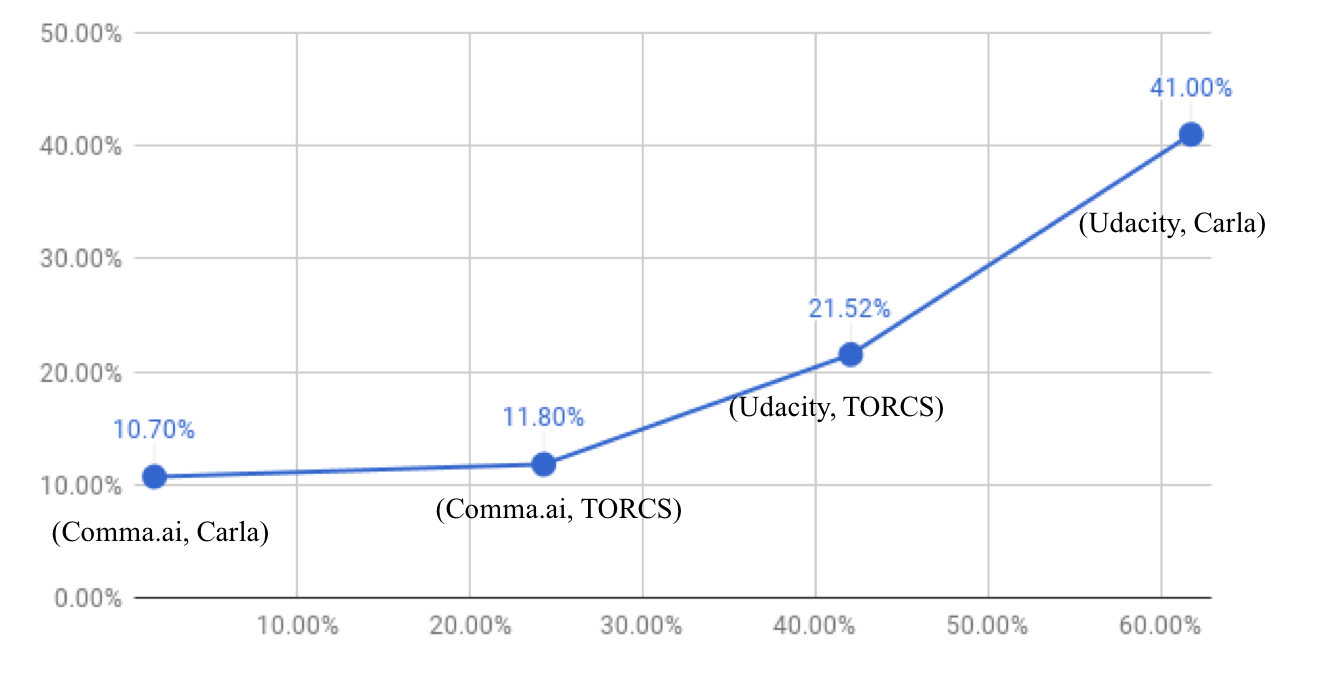}
	\caption{The percentage decrease in prediction MAE (y-axis) is positively correlated with the percentage decrease in input entropy (x-axis).}
	\label{entropy}
\end{figure}

Following the deduction in \ref{IB}, we now show that $H(X) > H(X_v)$, which infers $L_{\{T=X\}} > L_{\{T=X_v\}}$. While it is unclear how to measure the entropy of an arbitrary set of images, under the mild assumption of normal distribution, the entropy equals to the natural logarithm of the determinant of the covariance matrix up to a constant. We therefore treat each image as a vector and measure the total variance of 50 randomly sampled pairs of real and generated virtual data. As shown in Table \ref{var} and Fig \ref{entropy}, virtual representation tends to have lower entropy, giving a better initialization to the information bottleneck objective. The performance gain is positively correlated with the decrease in input entropy.

\begin{table}
	\centering
	\begin{tabular}{|c|c|c|c|c|}
		\hline
		\multirow{2}{*}{Variance} & \multicolumn{2}{c|}{Carla} & \multicolumn{2}{c|}{TORCS} \\ \cline{2-5} 
		& Udacity      & Commaai     & Udacity      & Commaai     \\ \hline
		Real                      & 82745        & 23902       & 107666       & 29656       \\ \hline
		Virtual                   & 31650        & 23483       & 62389        & 22453       \\ \hline
	\end{tabular}
	\caption{Variance of randomly sampled 50 pairs of real and generated virtual images. The generated virtual images have lower variance, which infers lower entropy for input distribution and thus less burden during the compression phase when optimizing the information bottleneck tradeoff.}
	\label{var}
\end{table}

\begin{figure}[ht!]
	\begin{center}
		\includegraphics[width=\linewidth]{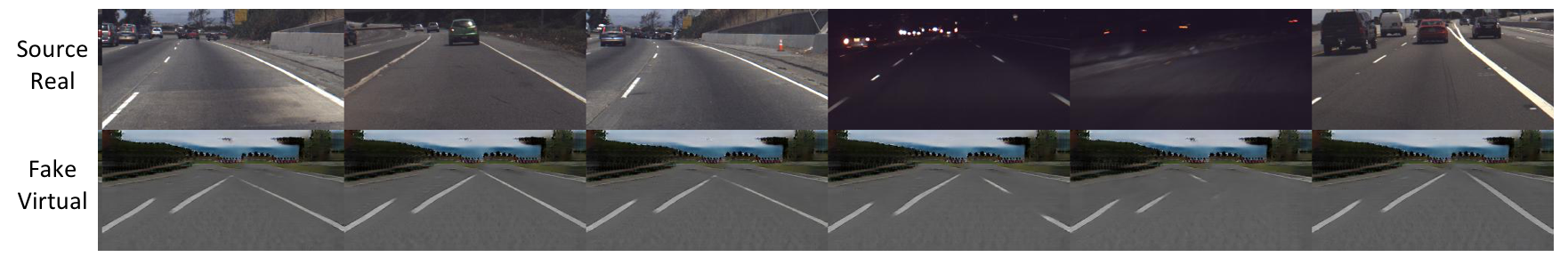}
	\end{center}
	\caption{Mode Collapse happens for naively implemented conditional GAN. }
	\label{fig:collapse}
\end{figure}
\subsection{Effectiveness of Domain Unification} 
\begin{figure}[ht]
	\begin{center}
		\includegraphics[width=\linewidth]{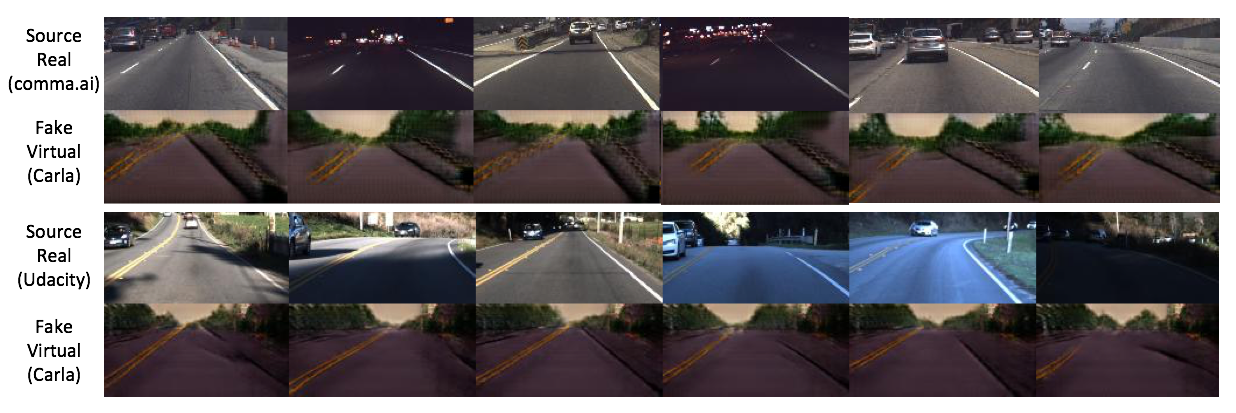}	
		\includegraphics[width=\linewidth]{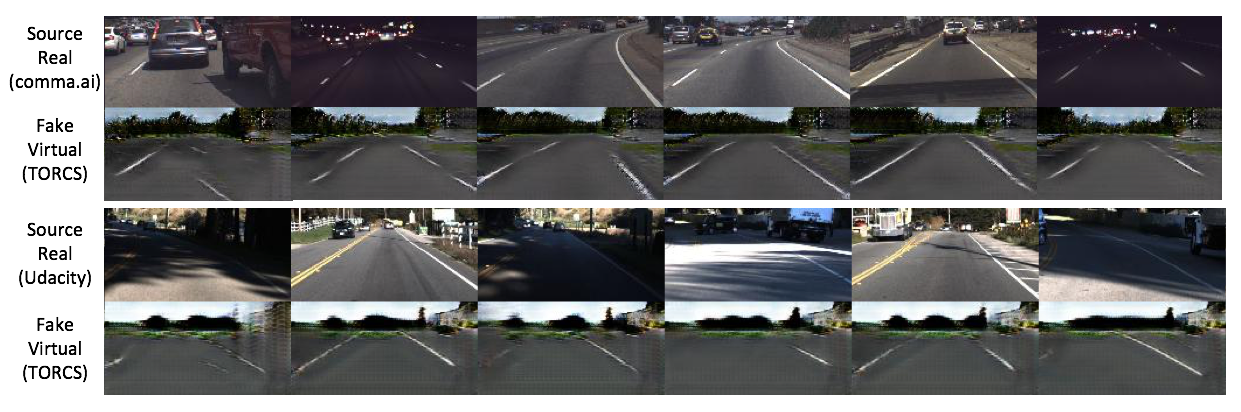}
	\end{center}
	\caption{Image generation results of DU-Drive. Information not critical to driving behavior, e.g. day/night lighting condition and the view beyond road boundary, is unified. Driving critical cues like lane markings are well preserved.}
	\label{fig:dudrive}
\end{figure}

A critical advantage of our model is that data collected from different sources could be unified to the same virtual domain. As shown in Figure \ref{fig:dudrive}, images from Comma.ai dataset and those from Udacity dataset are transformed into a unified virtual domain, whose superiority is directly reflected in the performance of steering angle prediction task. As shown in Table \ref{quant}, directly training a network with data from two real domains together will lead to results much worse than training each one separately due to domain shift. However, with DU-drive(unified), a single network could process data from multiple real domains with comparable results with DU-drive(single). Moreover, DU-drive separates the transformation and prediction process, and a generator could be independently trained for a new real dataset. 

To further study the generalization ability of DU-drive, we conducted semi-supervised experiments where labels are limited for an unseen dataset. We first train a DU-drive model with the Comma.ai data, then use 20\%/50\% of the labeled Udacity data respectively to train the generator with our co-training scheme and report the prediction performance on the test set. We also experimented on joint training with Comma.ai dataset under our domain unification framework. As shown in Table~\ref{semi}, Domain unification outperforms baselines by a large margin, especially when labeled data is scarce. This shows the superiority of domain unification at transferring knowledge across domains and alleviating domain shift.

\begin{table}
	\centering
	\begin{tabular}{|c|c|c|c|c|c|c|}
		\hline
		\multirow{2}{*}{\% of data used}               & \multicolumn{3}{c|}{Carla}   & \multicolumn{3}{c|}{TORCS}      \\ \cline{2-7} 
		& PilotNet & Ours(single) & Ours(unified) & PilotNet & Ours(single) & Ours(unified) \\ \hline
		20\%           & 7.86     & 7.12             & \textbf{6.02}         & 7.86     & 6.85             & \textbf{6.34}             \\ \hline
		50\%             & 7.11     & 6.41             & \textbf{5.15}       & 7.11           & 5.73      & \textbf{5.42}      \\ \hline
		100\%           & 6.02     & \textbf{3.57}           & 3.81      & 6.02     & 4.56             & \textbf{4.52}                \\ \hline
	\end{tabular}
	\caption{MAE for semi-supervised learning.}\label{semi}
\end{table}

\subsection{Prevention of mode collapse}
Mode collapse is a common pitfall for generative adversarial networks. Due to the lack of additional supervision, a naive implementation of conditional GAN easily suffers from unstable training and mode collapse (Figure \ref{fig:collapse}). With our novel joint training of steering angle prediction and real-to-virtual transformation, mode collapse for driving critical information like lane markings is effectively prevented.

\section{Conclusion}

We propose a real to virtual domain unification framework for autonomous driving, or DU-drive, that employs a conditional generative adversarial network to transform real driving images to their simpler counterparts in the virtual domain, from which vehicle control commands are predicted. In the case where there are multiple real datasets, a real-to-virtual generator could be independently trained for each real domain and a global predictor could be trained with data from multiple sources simultaneously. Qualitative and quantitative experiment results show that our model can effectively unify real images from different sources to more efficient representations in the virtual domain, eliminate domain shift and boost the performance of control command prediction task.
%
%
%
%

\end{document}